\documentclass[conference,a4paper]{IEEEtran}
\ifCLASSINFOpdf
\else
\fi
\usepackage[cmex10]{amsmath}
\usepackage{graphicx} 
\usepackage{booktabs}

\usepackage{amsmath}
\usepackage{amssymb}
\usepackage{amsfonts}
\usepackage{comment}
\usepackage{bm}
\usepackage{cite}

\begin{document}
%
\title{Memory-Efficient Point Cloud Registration via Overlapping Region Sampling}



%
\author{\IEEEauthorblockN{Tomoyasu Shimada,
Kazuhiko Murasaki,
Shogo Sato,
Toshihiko Nishimura,
Taiga Yoshida,
Ryuichi Tanida}

\IEEEauthorblockA{NTT Corporation, Japan}}


\maketitle
\newcommand{\boldsubsection}[1]{
  \noindent\textbf{#1}  
}

\begin{abstract}
Recent advances in deep learning have improved 3D point cloud registration but increased graphics processing unit (GPU) memory usage, often requiring preliminary sampling that reduces accuracy.
We propose an overlapping region sampling method to reduce memory usage while maintaining accuracy.
Our approach estimates the overlapping region and intensively samples from it, using a k-nearest-neighbor (kNN) based point compression mechanism with multi layer perceptron (MLP) and transformer architectures.
Evaluations on 3DMatch and 3DLoMatch datasets show our method outperforms other sampling methods in registration recall, especially at lower GPU memory levels.
For 3DMatch, we achieve 94\% recall with 33\% reduced memory usage, with greater advantages in 3DLoMatch.
Our method enables efficient large-scale point cloud registration in resource-constrained environments, maintaining high accuracy while significantly reducing memory requirements.
This paper is accepted for IEEE International Conference on Visual Communications and Image Processing 2024\footnote{\copyright 2024 IEEE. Personal use of this material is permitted. Permission from IEEE must be obtained for all other uses, in any current or future media, including reprinting/republishing this material for advertising or promotional purposes, creating new collective works, for resale or redistribution to servers or lists, or reuse of any copyrighted component of this work in other works.}.
\end{abstract}

\IEEEpeerreviewmaketitle    
\section{Introduction}

3D point cloud registration, the process of aligning multiple 3D datasets into a common coordinate system, is crucial in various applications such as autonomous driving, robotics, and augmented reality. 
Recent advances in deep learning-based methods have significantly improved the accuracy of registration.
However, these improvements often come at the cost of increased GPU memory usage, posing challenges for practical deployment, especially when dealing with large-scale point clouds.
To address such high memory consumption, pre-sampling techniques are commonly employed to reduce the number of points processed.
However, this approach can lead to a loss of critical geometric information, potentially compromising registration accuracy. 
The challenge lies in balancing computational efficiency with the preservation of essential point cloud information.

Our method introduces an approach that focuses on sampling from overlapping regions between point clouds. 
This method allows for efficient processing with reduced point counts, leading to decreased GPU memory usage without sacrificing the accuracy of subsequent registration steps.
By sampling from overlapping regions, we ensure that the most relevant information for registration is retained, even as the overall point count is reduced.
The key contributions of our work are:
\begin{itemize}
\item A new sampling strategy that prioritizes overlapping regions between point clouds.
\item Significant reduction in GPU memory usage through efficient point selection.
\item Maintenance of registration accuracy despite working with fewer points.
\end{itemize}

\begin{figure}
    \hspace*{20pt}
    \centering
    \includegraphics[width=0.9\linewidth]{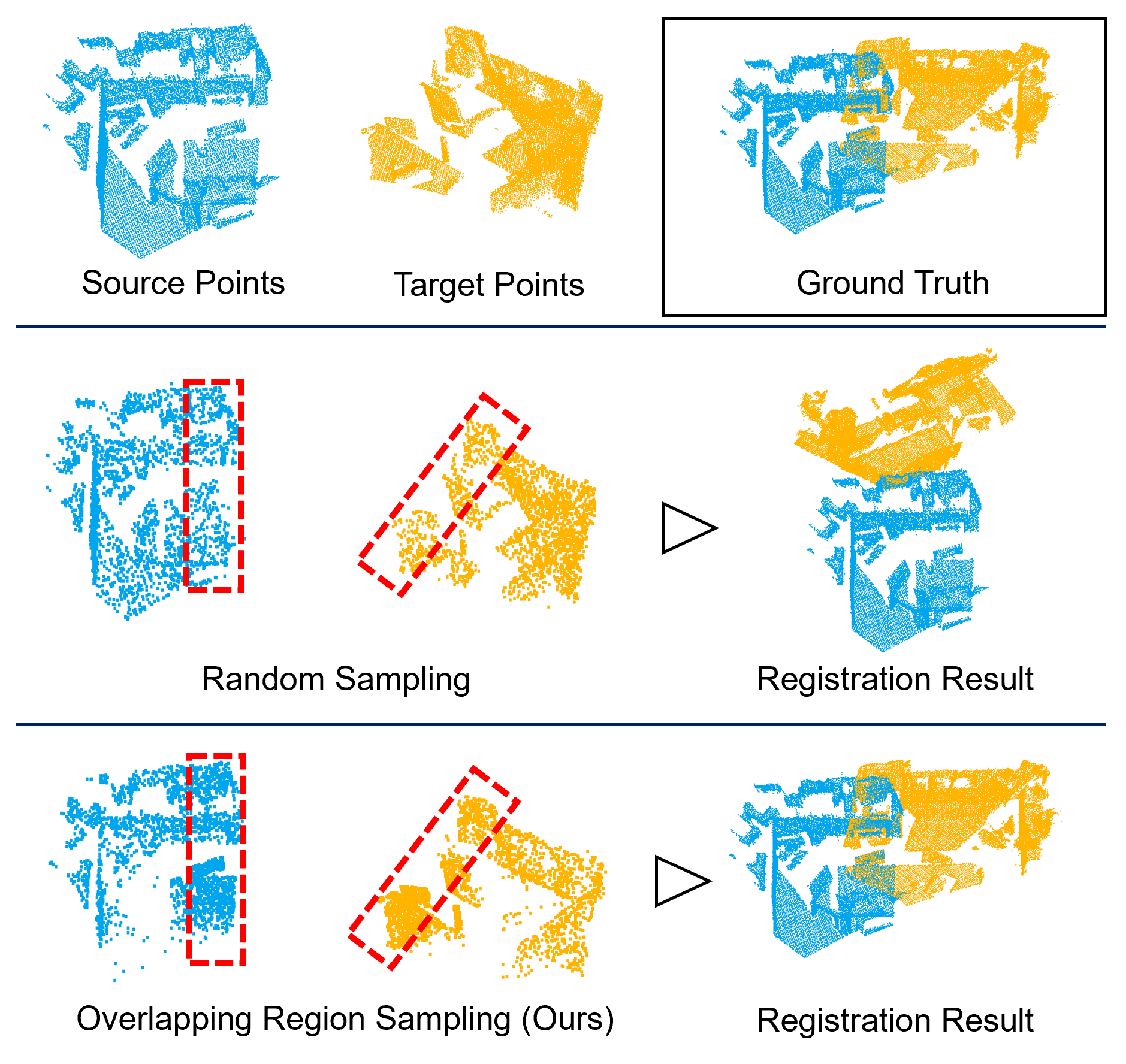}
    \caption{
    Comparison of point cloud registration using random sampling versus our overlapping region sampling method. Top: original source and target clouds with ground truth alignment. Red dashed boxes: true overlapping regions. Both methods reduce points to 20\% of original count. Middle: random sampling uniformly reduces points, potentially missing crucial overlapping areas. Bottom: our method concentrates points in overlapping regions (red dashed boxes) while reducing non-overlapping areas. Our sampling leads to more accurate registration, closely matching ground truth, despite using same number of sampled points.}
    \label{fig:intro}
\end{figure}

Fig.~\ref{fig:intro} demonstrates the efficiency of our method compared to random sampling for point cloud registration. 
Our approach preserves critical geometric information, particularly in overlapping regions, while reducing the number of input points. 
This visual comparison highlights our method's ability to achieve high-quality registration with significantly reduced GPU memory usage, addressing a key challenge in large-scale point cloud processing.

\begin{figure*}[t]
    \centering
    \includegraphics[width=0.85\linewidth]{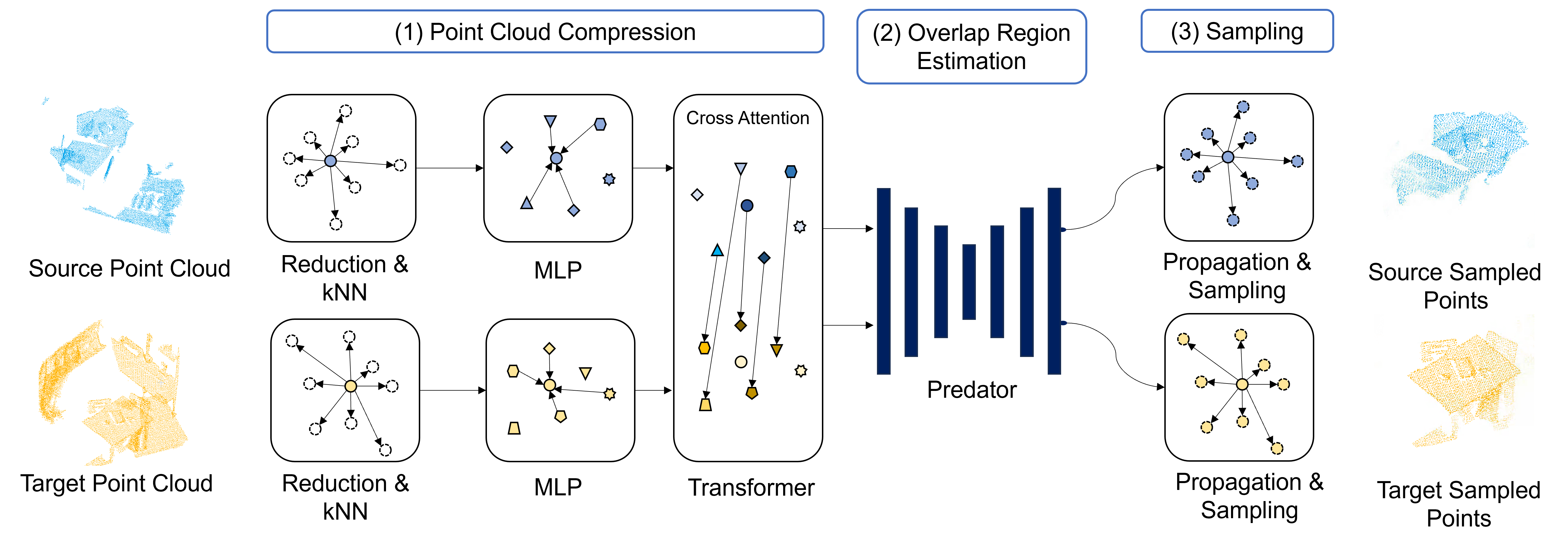}
    \caption{Overview of our proposed method to sample from overlap region for efficient 3D point cloud registration. The process consists of three main stages: (1) Point Compression, which includes point reduction, kNN-based feature extraction, and cross attention between source and target point clouds; (2) Overlap Region Estimation using the PREDATOR framework; and (3) Sampling, which involves score propagation and overlap-aware point selection. This approach enables memory-efficient processing while preserving crucial geometric information for accurate registration.}
    \label{fig:knn_graph}
\end{figure*}

\section{Related Work}
\label{sec:relate}
3D point cloud registration is a fundamental problem in 3D computer vision with applications ranging from autonomous driving to augmented reality. In this section, we first review existing point cloud registration methods, then discuss various point cloud sampling techniques that are crucial for efficient processing of large-scale point clouds.

\subsection{Point Cloud Registration}
Classical methods such as Iterative Closest Point (ICP)~\cite{icp} and its variants~\cite{goicp} iteratively estimate correspondences and transformations. While computationally efficient for small point clouds, these methods struggle with large-scale data and are sensitive to initial alignments. Recent advances include robust estimators \cite{ransac,sdrsac} and global optimization techniques \cite{mac}, which improve accuracy but at a higher computational cost. For example, TEASER++ \cite{teaser} achieves state-of-the-art robustness, but requires $O(N^2)$ memory for $N$ points, limiting its scalability.


Recent works focus on learning robust feature descriptors and matching correspondences~\cite{perfect, predator, ppfnet, cofinet, dcm, fcgf, lepard, roitr, spin, d3feat}. 
FCGF~\cite{fcgf} introduces fully convolutional geometric features, while D3Feat~\cite{d3feat} proposes a joint learning framework for detection and description.
Transformer-based architectures like PREDATOR \cite{predator} and GeoTransformer~\cite{geotransformer} have further improved performance by capturing long-range contextual information.
PREDATOR reports a registration recall of 89.0\% on 3DMatch, outperforming previous methods by a significant margin.

Although these deep learning techniques show promise in terms of accuracy and robustness, their GPU memory usage scales with the number of points in the cloud, posing challenges for processing large-scale point clouds.
This limitation motivates the need for effective point cloud sampling methods.

\subsection{Point Cloud Sampling Methods}

Point cloud sampling is a crucial preprocessing step in 3D registration tasks, aiming to reduce computational complexity while preserving important geometric information. 
Several sampling strategies have been proposed and widely used in the literature.

Random sampling is one of the simplest and most straightforward methods for point cloud reduction.
It randomly selects a subset of points from the original point cloud. While computationally efficient, it may not preserve the geometric structure of the original point cloud effectively.
Poisson disk sampling~\cite{pds} generates a uniform distribution of points while maintaining a minimum distance between any two points.
This method is effective in preserving the overall shape of the point cloud while reducing the number of points.
Voxel grid sampling~\cite{vgs} divides the point cloud space into a 3D grid of voxels.
For each occupied voxel, it replaces all points within the voxel with their centroid.
This method effectively reduces the point cloud density while maintaining a uniform spatial distribution.
Farthest point sampling~\cite{fps1,fps2} iteratively selects the point that is farthest from the already selected points.
This method ensures a good coverage of the original point cloud geometry and is particularly effective in preserving global structures.

Each of these sampling methods has its own strengths and weaknesses in terms of computational efficiency, geometric preservation, and distribution uniformity.
The choice of sampling method can significantly impact the subsequent registration process, affecting both accuracy and computational requirements.
Our proposed method aims to address the limitations of existing approaches by introducing a sampling technique that maintains registration accuracy while significantly reducing memory requirements, thus enabling efficient processing of large-scale point clouds.
\section{Overlapping Region Sampling}
\label{sec:method}

Fig.~\ref{fig:knn_graph} illustrates an overview of our proposed method for efficient point cloud registration through intelligent sampling.
Our approach takes a pair of point clouds as input and learns to identify overlapping regions using supervised learning.
The method consists of three key stages: (1) Point Cloud Compression, (2) Overlap Region Estimation, and (3) Sampling.
By focusing on the most relevant points in the overlapping areas, our method significantly reduces computational requirements while maintaining registration accuracy.
As shown in Fig.~\ref{fig:knn_graph}, our method estimates overlapping regions between pairwise point clouds using supervised learning.
In the following, the process of extracting the point clouds \(\hat{P}\) and \(\hat{Q}\) based on the estimation of overlapping regions will be explained, using the source point cloud \( P = \{\bm{p}_i \in \mathbb{R}^3\mid i = 1 \ldots n_p \} \) and the target point cloud \( Q = \{\bm{q}_i \in \mathbb{R}^3\mid i = 1 \ldots n_q \} \) as inputs.

\subsection{Point Cloud Compression with kNN}
We propose a two-stage approach for efficient point cloud compression that preserves geometric information while reducing GPU memory usage.
Our method combines local feature extraction with global context integration for comprehensive point cloud representation.

The first stage involves random sampling of representative points and kNN-based feature extraction.
We employ a simple Multi-Layer Perceptron (MLP) to encode local geometric context from kNN.
We denote random sampled points as $\tilde{P}$ and $\tilde{Q}$, and the encoded features as $\tilde{F}^P$ and $\tilde{F}^Q$.
This MLP efficiently captures local structures in a compact representation, maintaining crucial geometric relationships despite significant point reduction.
In the second stage, we utilize Transformer~\cite{transformer} for cross-attention between compressed features of paired point clouds.
This enhances feature representations by incorporating global context and paired point cloud relationships.
Transformer~\cite{transformer} for point cloud $P$ uses features $\tilde{F}^P$ as the query and $\tilde{F}^Q$ as the key and value, producing weighted features $\tilde{F}^{P \leftarrow Q}$ that encapsulate both local and global information.


\subsection{Overlap Region Estimation}

To estimate overlapping regions, we adapt the PREDATOR framework, which is designed to handle point cloud pairs with low overlap.
We modify PREDATOR to use our compressed and enhanced features $\tilde{F}^{P \leftarrow Q}, \tilde{F}^{Q \leftarrow P}$ as input.
The PREDATOR module processes these features and outputs two key scores for each point in the sampled point clouds: an overlap score $O$ and a matchability score $M$.
The overlap score represents the probability of a point being in the overlapping region between the two point clouds, while the matchability score indicates the likelihood of matching in the feature space with the corresponding point of the other point cloud.
To identify points that are both in the overlapping region and likely to have good correspondences, we combine these scores by taking their product.
We define the combined score is $s_{\bm{p}_i} = o_{\bm{p}_i} \times m_{\bm{p}_i}$, 
where $o_{\bm{p}_i}$ and $m_{\bm{p}_i}$ are the overlap and matchability scores for the point $\bm{p}_i$, respectively.

This combined score effectively highlights points that are crucial for accurate registration, as they are both in the overlapping region and likely to be matchable.
The resulting combined scores provide a point-wise measure of importance for the registration process, allowing us to focus on the most relevant points in subsequent steps.
This approach is particularly effective in scenarios with partial overlaps or in the presence of noise, where identifying reliable correspondences is critical for accurate registration.

\begin{figure*}[t]
    \centering
    \includegraphics[width=0.85\linewidth]{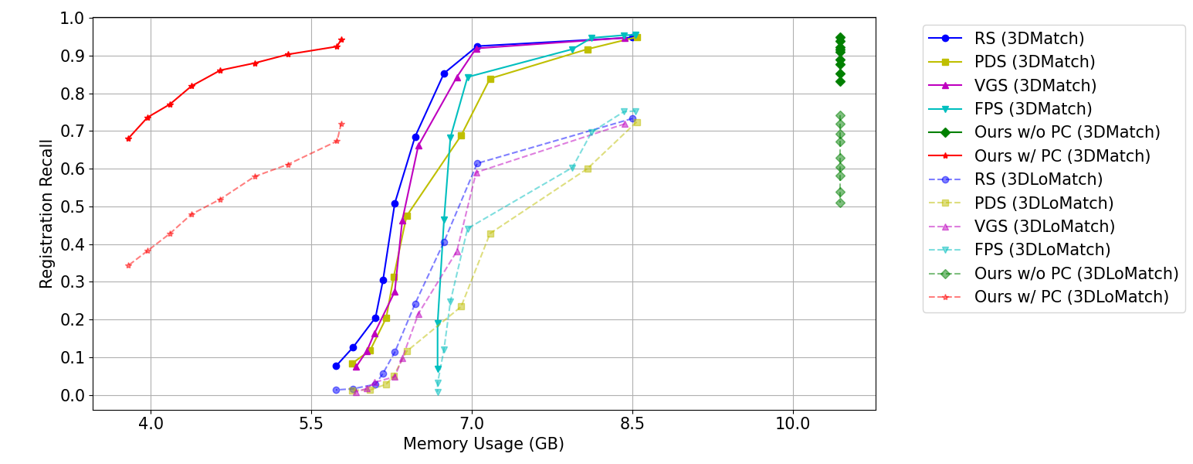}
    \caption{Comparison of registration recall versus memory usage for different point cloud sampling methods, all using GeoTransformer for registration, on the 3DMatch and 3DLoMatch datasets. The graph compares various sampling techniques including random sampling (RS), Poisson disk sampling (PDS), voxel grid sampling (VGS), and farthest point sampling (FPS), as well as the proposed method with and without points compression (PC). Solid lines represent performance on 3DMatch, while dashed lines show results for the more challenging 3DLoMatch dataset. The proposed sampling method ('Ours') consistently achieves higher recall rates with lower memory usage across both datasets, particularly when using point compression.}
    \label{fig:recall_vs_memory}
\end{figure*}

\subsection{Loss Function}

We employ a composite loss function to train our overlap region estimator, adapting the approach from PREDATOR~\cite{predator}. 
$\omega_c$, $\omega_o$, and $\omega_m$ are weighting factors.
The total loss is a weighted sum of three components $
    \mathcal{L} = \omega_c\mathcal{L}_c + \omega_o\mathcal{L}_o + \omega_m\mathcal{L}_m$.
Each component is computed for $P$ and $Q$, respectively (i.e. $\mathcal{L}_i = \frac{1}{2}(\mathcal{L}_i^P+\mathcal{L}_i^Q) , \quad i \in \{c, o, m\}$).
First, the circle loss $\mathcal{L}_c$ supervises the feature descriptors, encouraging similarity between corresponding points and dissimilarity between non-corresponding points. 
Specifically, it is calculated as follows:

\vspace{-0.5\baselineskip}
\begin{equation}
\scalebox{0.75}{$\displaystyle
\mathcal{L}_c^P = \frac{1}{|\tilde{P}|} \sum_{\bm{p}_i \in \tilde{P}} \log \left[ 1 + \sum_{j \in \varepsilon_c (\bm{p}_i)} e^{\beta_c (d_{i,j} - \Delta_c)} \cdot \sum_{k \in \varepsilon_n (\bm{p}_i)} e^{\beta_n (\Delta_n - d_{i,k})} \right] ,
$}
\end{equation}
\vspace{-0.5\baselineskip}

where $\varepsilon_c (\bm{p}_i)$ is corresponding points of $\bm{p}_i$ in $\tilde{Q}$ and $\varepsilon_n (\bm{p}_i)$ is non-corresponding points, $d_{i,j}$ represents the distance between $\bm{p}_i$ and $\bm{q}_j$ in feature space, $\beta_c, \beta_n, \Delta_c$ and $\Delta_n$ are hyperparameters.
Next, the overlap loss $\mathcal{L}_o$ is implemented as a binary cross-entropy loss for overlap probability estimation.
For point cloud $P$, it is defined as:

\vspace{-0.5\baselineskip}
\begin{equation}
\scalebox{0.75}{$\displaystyle
    \mathcal{L}_o^P = -\frac{1}{|\tilde{P}|} \sum_{\bm{p}_i \in \tilde{P}} [y_{\bm{p}_i} \log(o_{\bm{p}_i}) + (1 - y_{\bm{p}_i}) \log(1 - o_{\bm{p}_i})],
$}
\end{equation}
\vspace{-0.5\baselineskip}

where $y_{\bm{p}_i}$ is the ground truth overlap label and $o_{\bm{p}_i}$ is the estimated overlap probability for point $\bm{p}_i$ in $\tilde{P}$. 
Finally, the matchability loss $\mathcal{L}_m$ encourages the network to predict matchable points. It is defined similarly to the overlap loss, but uses matchability ground truth labels.
For point cloud $P$, it is defined as:

\vspace{-0.5\baselineskip}
\begin{equation}
\scalebox{0.75}{$\displaystyle
    \mathcal{L}_m^P = -\frac{1}{|\tilde{P}|} \sum_{\bm{p}_i \in \tilde{P}} [\bar{m}_{\bm{p}_i} \log(m_{\bm{p}_i}) + (1 - \bar{m}_{\bm{p}_i}) \log(1 - m_{\bm{p}_i})],
    $}
\end{equation}
\vspace{-0.5\baselineskip}

where $\bar{m}_{\bm{p}_i}$ is the ground truth matchability label for point $\bm{p}_i$ in $\tilde{P}$.
$\bar{m}_{\bm{p}_i}$ becomes $1$ when the distance of real space between the transformed point of $\bm{p}_i$ by true transformation parameter and the nearest neighbor point belonging $\tilde{Q}$ in the feature space from $p_i$ is less than the threshold $r_m$, and otherwise $0$.

\subsection{Score Propagation and Overlap-aware Sampling}

After obtaining the overlap scores $S$ for the sampled points $\tilde{P}, \tilde{Q}$, we propagate these scores to the original, unsampled points in the point clouds $P, Q$.
This propagation ensures overlap information for all points, including those initially removed during compression.
For each unsampled point, we identify its five nearest neighbors among the sampled points and assign the average of their overlap scores.
The hyperparameter five is determined experimentally.

Using these propagated overlap probabilities, we perform a final sampling step to focus on the most relevant points for registration.
We select a subset of points from each point cloud that have the highest overlap probability values.
This overlap-aware sampling enables efficient registration with reduced memory usage while retaining the most important information for accurate alignment.
The resulting set of points, enriched with overlap information, is then used for the subsequent registration step.

\section{Experiments}
\label{sec:experiments}

\subsection{Implementation and Training}
We implemented our method using PyTorch and trained it on a single NVIDIA A100 GPU.
The model was trained for 40 epochs with a batch size of 1, optimizing parameters based on performance.
In our point cloud compression module, we set the sampling rate to one-fifth of the input point count, striking a balance between memory efficiency and preservation of geometric information.
During initial training epochs, we set $\omega_c$ and $\omega_o$ to 1, while $\omega_m$ was set to 0.
Once the accuracy of interest point matching reached 30\%, we adjusted $\omega_m$ to 1.
We used the pre-processed 3DMatch dataset~\cite{3dmatch} provided by~\cite{predator} for training.
Other hyperparameters, including $\Delta_c$, $\Delta_n$, $\beta_c$, $\beta_n$ and $r_m$, were set according to PREDATOR~\cite{predator}.
To evaluate registration accuracy, we employed GeoTransformer~\cite{geotransformer}.

\subsection{Evaluation}

\boldsubsection{Overlap Region Estimation Accuracy:}We conducted an ablation study to evaluate our point cloud compression (in Fig.~\ref{fig:knn_graph}  (1)) effectiveness in extracting overlapping regions while reducing memory usage.
We compared our method with PREDATOR, which serves as a baseline without the point cloud compression.
Table \ref{table:ablation} shows the results.
\begin{table}[t]
\renewcommand{\arraystretch}{1.3}
\centering
\small
\caption{Effect of point cloud compression on overlapping region extraction and memory usage}
\begin{tabular}{c|ccc}
\hline
\rule{0pt}{2.5ex} 
Point Cloud & GPU Usage & \multicolumn{2}{c}{AP} \\
Compression& (GB) & 3DMatch & 3DLoMatch \\
\hline
\rule{0pt}{2.5ex}
w/  & 3.70 & 0.852 & 0.600 \\
w/o  & 10.44 & 0.884 & 0.647 \\
\hline
\end{tabular}

\label{table:ablation}
\end{table}
Our module reduces memory usage by about 64.6\% compared to the baseline PREDATOR method.
Despite these significant memory savings, we maintain competitive performance in overlapping region extraction as measured by Average Precision (AP).
In the 3DMatch data set, the AP of our method is only 3.2 points lower than the baseline.
For the more challenging 3DLoMatch dataset, the AP decrease is 4.7 points lower.

\boldsubsection{Registration Accuracy:}
We evaluated registration accuracy using various point cloud sampling methods, all employing GeoTransformer for the registration process, on both 3DMatch and 3DLoMatch datasets.
Fig.~\ref{fig:recall_vs_memory} shows the relationship between GPU memory usage and registration recall for different sampling techniques.

As shown in Fig.~\ref{fig:recall_vs_memory}, our proposed method consistently outperforms other sampling techniques, including random sampling (RS), Poisson disk sampling (PDS), voxel grid sampling (VGS), and farthest point sampling (FPS), in terms of registration recall at various levels of memory usage.
This performance advantage is observed in both the 3DMatch~\cite{3dmatch} and the more challenging 3DLoMatch~\cite{predator} datasets.
It is important to note that the GPU memory usage reported in Fig.~\ref{fig:recall_vs_memory} represents the maximum memory consumption observed during either the sampling or registration process, whichever is greater. 
For the 3DMatch dataset, our method maintains a high registration recall above 0.8 using 4.38 GB of GPU memory. 
Our proposed method is about 33\% more efficient in GPU usage than random sampling.
The absence of point cloud compression also shows high recall at all sampling rates, indicating the effectiveness of overlapping region sampling.
However, due to the large GPU memory required for sampling, the method with the point cloud compression module is about 80\% more memory efficient.


The results clearly demonstrate the effectiveness of our proposed approach in balancing registration accuracy and computational efficiency. 
By focusing on potential overlapping regions and utilizing point cloud compression, we maintain high registration recall while significantly reducing memory requirements. 
Our method's consistent performance across both datasets indicates robustness to varying degrees of overlap, a crucial feature for diverse real-world applications. 
This approach enables efficient large-scale point cloud registration without compromising accuracy, even in resource-constrained environments.

\section{Conclusion}
\label{sec:conclusion}

This paper presents a novel sampling method for large-scale 3D point cloud registration that significantly improves performance while minimizing GPU memory usage.
Our approach demonstrates superior registration recall compared to random sampling on both 3DMatch and 3DLoMatch datasets, achieving over 90\% recall with only 5 GB of GPU memory.
This efficiency enables robust processing of large-scale point clouds in resource-constrained environments.

Future work will focus on evaluating even larger and more diverse point cloud datasets, assessing performance on various hardware configurations, and further optimizing memory efficiency for extremely large point clouds. 
This contribution marks a significant advancement towards efficient and accurate large-scale 3D point cloud registration, opening new possibilities for processing extensive 3D data in real-world scenarios.


\bibliographystyle{IEEEtran}
\bibliography{main}

\end{document}